\documentclass[12pt, letterpaper]{article}
\usepackage[utf8]{inputenc}
\usepackage[english]{babel}
\usepackage{wrapfig}
\usepackage{bm}
\usepackage{amsmath}
\newcommand{\STM}{{\mathchoice{}{}{\scriptscriptstyle}{} STM}}
\newcommand{\PC}{{\mathchoice{}{}{\scriptscriptstyle}{} PC}}
\usepackage[pdftex]{graphicx}
\DeclareGraphicsExtensions{.pdf,.jpeg,.png}
\usepackage{authblk}
\usepackage{cite}

\title{Spoken digit classification using spin-wave delay-line active-ring reservoir computing}
\author{Stuart Watt}
\author{Mikhail Kostylev}
\affil{School of Physics and Astrophysics, University of Western Australia, Crawley, W.A. 6009, Australia}

\begin{document}
\maketitle{}

\begin{abstract}
As a test of general applicability, we use the recently proposed spin-wave delay-line active-ring reservoir computer to perform the spoken digit recognition task. On this, classification accuracies of up to 93$\%$ are achieved. The tested device prototype employs improved spin wave transducers (antennas). Therefore, in addition, we also let the computer complete the short-term memory (STM) and parity check (PC) tasks, because the fading memory and nonlinearity are essential to reservoir computing performance. The resulting STM and PC capacities reach maximum values of 4.77 and 1.47 respectively.
\end{abstract}

\section{Introduction}
The Reservoir Computer (RC) is a machine learning model based on the recursive neural network \cite{jaeger_harnessing_2004, verstraeten_experimental_2007} which has shown impressive suitability for analyzing complex dynamical systems. The simplicity of RC models have also been shown to allow hardware implementations leading to a growing field of research. The physical RC offers fast, powerful and energy efficient alternatives to traditional computing by exploiting the rich nonlinear dynamics available in natural phenomena. Among those so far put forward (see Ref. \cite{tanaka_recent_2019} for a review), spintronic based architectures are promising candidates for practical RC applications due to their low power usage, strong nonlinearity arising from magnetization dynamics, and their scalability.\par

In a recent publication we proposed an implementation of a physical RC using propagating spin waves circulating in an active-ring resonator \cite{watt_reservoir_2020}. That work showed how the auto-oscillatory and nonlinear phenomena of magnetic film active-ring resonators, which have been extensively studied over the past two decades \cite{wu_nonlinear_2010}, can be applied to this novel area of research.

In this letter, we continue to evaluate the suitability of the active-ring system as a RC. We employ the same Magnetostatic Surface Spin Wave (MSSW) based active ring, but with improved spin wave transducers. We demonstrate that the active-ring system is able to successfully complete the spoken digit recognition task. The dataset chosen for this task is based on the TI 46 corpus \cite{TI46}.  As a preliminary step, we also run the short-term memory (STM) task \cite{jaeger_tutorial_2002} and the parity check (PC) task \cite{furuta_macromagnetic_2018, bertschinger_real-time_2004} on the improved device. \par

\section{Experimental Methods}
\begin{figure*}[!t]
	\centering
	\includegraphics[width=1\textwidth]{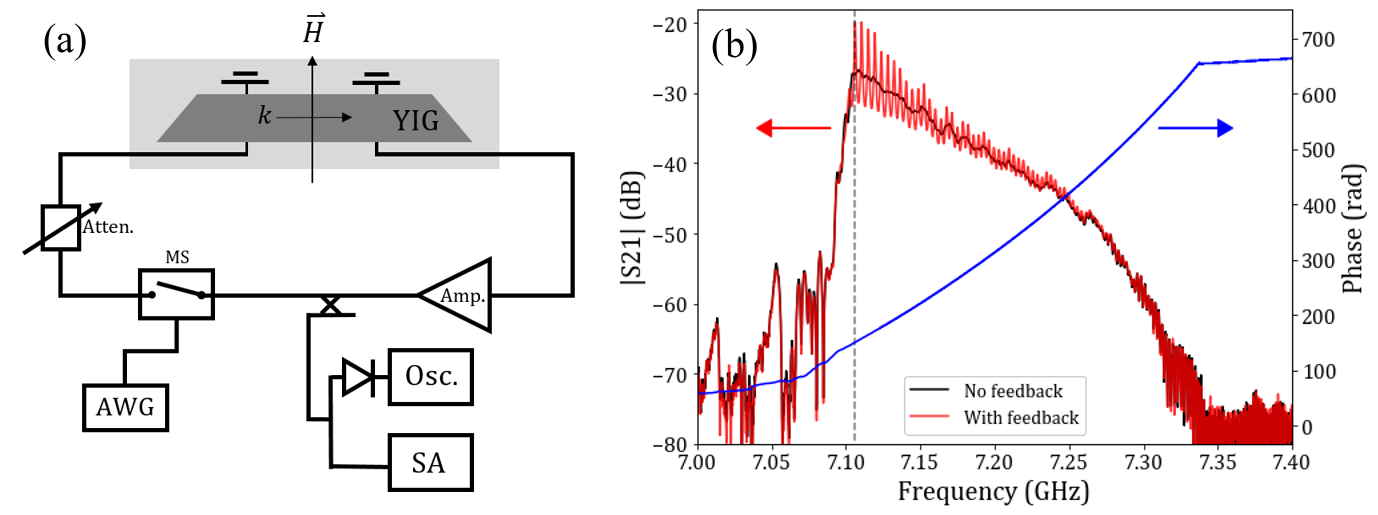}
	\hfil
	\caption{(a) Schematic diagram of the spin-wave delay-line active-ring resonator system. (b) Transmission amplitude and phase characteristic of the spin-wave delay line. The trace in black shows the transmission band of the delay line alone. When the feedback loop is closed and the feedback gain increased, ring eigenmodes are formed when the spin wave wavenumber satisfies the condition $kd=2 \pi m$, where $d$ is the antenna separation and $m$ is a positive integer. The vertical dashed line shows the eigenfrequency with the lowest loss, which is the first auto-oscillation frequency. In this case $f_{res}=7.1$ GHz. The magnetic field is 1700 Oe.}
	\label{fig:FIG1}
\end{figure*}

The experimental setup depicted schematically in Fig. \ref{fig:FIG1}(a) consists of a spin-wave delay-line active-ring resonator, with a microwave PIN diode switch inserted into the feedback loop to allow fast voltage control of the feedback gain. For the theory of operation of this active ring as a RC implementation we refer the reader to Ref. \cite{watt_reservoir_2020}, however some enhancements on the original design concept should be noted here. \par

Firstly, the delay-line structure in the present work makes use of thinner microstrip antennas (50 $\mu$m in the resent work compared to 0.5 mm in previous works) which drastically increases the efficiency of spin wave excitation as well as broadening the transmission band (see Fig. \ref{fig:FIG1}(b)). The latter effect would be useful when linking multiple delay lines in series to increase the complexity of the RC. \par

Another improvement is the use of a thinner YIG film. For the MSSW excited in this active ring, the spin wave group velocity decreases along with the YIG film thickness, and hence the delay time has been increased. In the present work a 5.5 $\mu$m thick and 2 mm wide YIG strip is employed as the delay line. With an antenna separation of 8.3 mm, the signal circulation time, $T_{r}$, of the active ring has been increased from approximately 50 ns in our previous work to 236 ns.\par

\section{Performance evaluation on benchmark tests}
\subsection{Short-term memory and Parity-check tasks}

An effective physical RC implementation must satisfy two main properties: a fading memory to allow processing of time dependent signals and nonlinear dynamics to allow for linear separability of different inputs. These properties can be assessed using the STM and PC tasks respectively. For the specific details on how the STM and PC tasks are defined as well as how they are carried out on our spin-wave delay line active ring RC, the reader is referred to Ref. \cite{watt_reservoir_2020}, however brief details are given below.\par

\begin{figure*}[!t]
	\centering
	\includegraphics[width=1\textwidth]{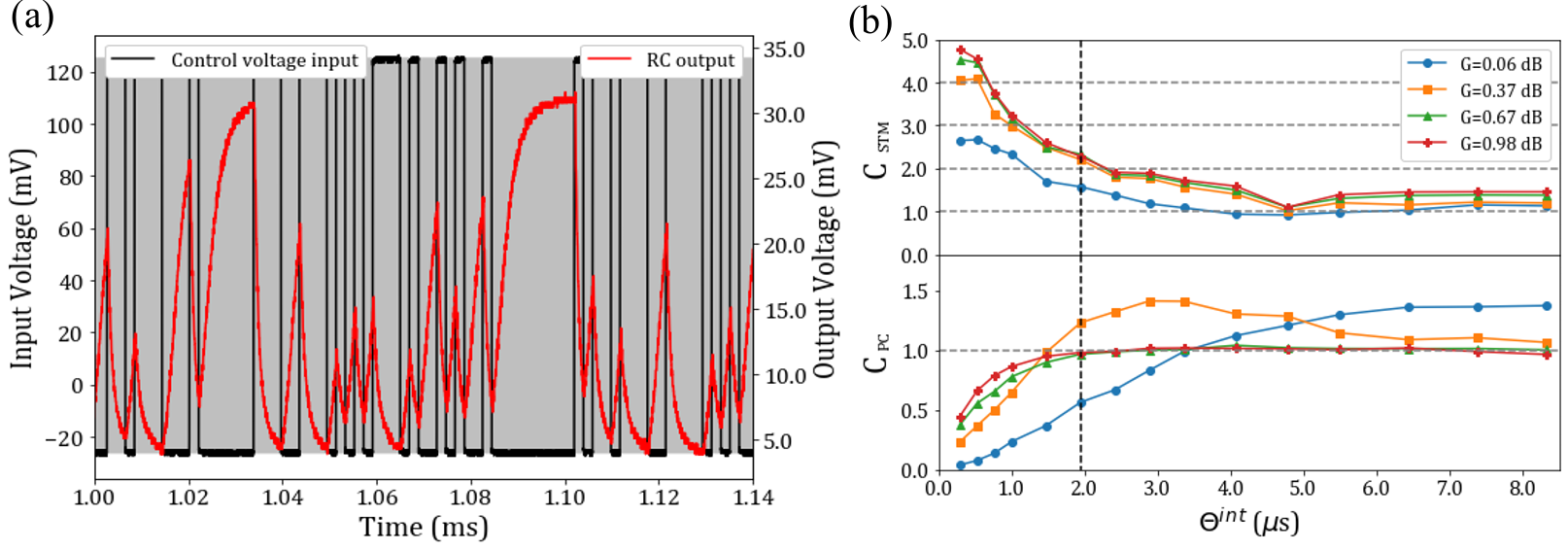}
	\hfil
	\caption{(a) Example of the RC input voltage (black) and the active-ring diode output voltage (red). The grey shaded region indicates the computing range of input control voltage. The input interval time is $\theta^{int}=1.95$ $\mu$s. (b) Delay-line active-ring RC performance on STM (top) and PC (bottom) tasks as a function of the input time interval $\theta^{int}$ at various feedback gain $G$. The vertical dashed line shows the input time interval $\theta^{int}=1.95$ $\mu$s.}
	\label{fig:FIG2}
\end{figure*}

To implement these tests, a binary series of 2200 values is fed sequentially into the RC by way of a voltage applied by an arbitrary waveform generator to the microwave switch (‘AWG’ and ‘MS’ respectively in Fig. \ref{fig:FIG1}(a)). The binary values of 0 and 1 correspond to the boundary values of the input control voltage range $V_{in}$=[-25 mV,125 mV]. Within this voltage region the microwave switch provides an approximately linear attenuation. Each input voltage is applied to the switch for a fixed interval of time which has length $\theta^{int}$. A fast microwave diode is used to measure the amplitude of the signal in the active ring over this time interval to create the RC output. Fig. \ref{fig:FIG2}(a) shows the binary series input voltage applied to the microwave switch and the active-ring diode output.\par

The delay-line active-ring resonator system acts essentially as a single neuron with feedback onto itself. In order to increase the dimensionality  of the system we employ the technique known as time-multiplexing \cite{appeltant_information_2011}. In this method, each input is injected into a reservoir of $N_{\theta}$ ‘virtual’ neurons, separated in time and connected in series. In this way one may view a single dynamical node as an entire reservoir of ‘virtual’ neurons. \par

For each time interval of length $\theta^{int}$, the active ring signal is sampled a total of 20 times to create an output vector. This vector represents the output from a reservoir with $N_{\theta}=20$  ‘virtual’ neurons. A linear regression model is then trained on these 20 neuron values to try and recreate the ‘target’. The targets are what the RC model is trying to predict and depend on the specific task at hand. \par

The purpose of the RC is to process the original binary series into a format from which the linear regression can more easily reconstruct the desired target. To this end, the ‘fading’ memory enabled by the slow transient process of the microwave signal in the active ring means that the outputs for each time interval have influence from multiple previous inputs as well as the current input. The transient process of establishing a stable auto-oscillation amplitude in the ring is enabled by the presence of nonlinear damping of the spin waves. The nonlinear damping also helps to linearly separate the different inputs.\par

The entire 2200 value binary series is fed into the active ring and 2200 output vectors are recorded. The first 200 outputs are discarded to wash out any transient signal behavior in the RC which may depend on the initial conditions of the active ring. The following 1000 outputs are used for training the linear regression and the remaining 1000 outputs are used to evaluate the performance of the model. The STM task is a memory recall task, where the targets are defined as the input values at some time delay in the past. For the PC task, the targets are the result of a binary parity operation over previous inputs up to some delay in the past. In both cases, the targets are themselves binary values. Since the targets depend on the amount of delay between the input and the output, a different linear regression model is trained for each level of delay.\par

The success of the linear regression to reconstruct the desired target is measured by calculating the square of the correlation coefficient between the reconstructed target and the actual target. This is done for the various delays and the results are summed together to obtain the STM and PC capacities ($C_{\STM}$ and $C_{\PC}$ respectively) shown in Fig. \ref{fig:FIG2}(b). These plots show the short term memory and the nonlinear performance as a function of the input time interval $\theta^{int}$ and the feedback gain strength $G$.\par

The results are qualitatively similar to those presented in Ref. \cite{watt_reservoir_2020}. The STM capacity is maximized for short input time intervals, consistent with the notion that the system cannot reach equilibrium fast enough and thus current inputs influence successive outputs. The PC capacity reaches a peak at an intermediate input time interval which decreases as the feedback gain is raised. The main difference of these results to those published already are the quantitative improvement of the values. The maximum STM capacity obtained in this work is 4.77 for $\theta^{int}=0.295$ $\mu$s and a high feedback gain. The maximum PC capacity is 1.47, which is obtained for an intermediate value of $\theta^{int}$ and for lower gains closer to the threshold. The improvements on these benchmark tests from our previous work can be attributed to the modifications made to the experimental setup mentioned above.\par

With the lower group velocity and longer delay time associated with the thinner YIG film, the system reacts much slower to a given input. We found that the STM and PC capacities are quite sensitive to the synchronization between the inputs and the round trip time of the active ring, with optimal results occurring for input time intervals which are equal to $\theta^{int}=(m+1/4)T_{r}$ where $m$ is an integer. The data points in Fig. \ref{fig:FIG2}(b) satisfy this condition. Further details of this result will be published elsewhere.\par

\subsection{Spoken digit recognition task}
The STM and PC tasks are simple metrics which allow easy RC performance measures and comparison between different physical RC concepts. To show the use of our RC concept in a more practical application, we perform the commonly used spoken digit recognition task. The data used task is a subset of the TI 46 corpus \cite{TI46}. This dataset involves audio samples of the 10 digits, each uttered 10 times by 5 different female speakers to make a total of 500 digit samples. These audio signals have been sampled at a rate of 12.5 kHz and have varying lengths. \par

Before injection into the experimental setup, appropriate pre-processing methods must be applied to the dataset to convert the waveforms into formats manageable by the RC. Typical speech recognition techniques analyze waveforms in the frequency domain and there are varying ways of doing this. Ref. \cite{abreu_araujo_role_2020} shows how nonlinear operations involved in the pre-processing stage can themselves result in adequate digit classification accuracy even without using a reservoir. Since the purpose of performing this task is to evaluate the capability of the spin-wave delay-line active ring to act as a RC, we purposefully choose a pre-processing method which is devoid of any nonlinear operations. Thus any nonlinearity comes from our experimental setup. To this end, we select a simple spectrographic approach to processing the dataset. \par

To being with, each digit waveform is broken into individual sound units of uniform duration $\tau$. Training of the models and classification is performed on these individual sound units instead of over the entire waveform to mimic the way spoken language processing treats speech signals as a sequence of phonemes. Given that the temporal characteristics of a speech signal can change rapidly, selecting sufficiently short unit time intervals ensures stationarity of the signal across that unit. We employ a 20 ms duration interval, resulting in a vector of 250 data points for each unit. Each digit in the database has varying length and so is segmented into a different number $N_{int}^{j}$ of sound unit intervals. Here the superscript $j$ references the specific digit sample used from the dataset. A fast Fourier transform operation is applied to each sound unit and only the real components of the Fourier image are kept. The resulting column vector $\vec{k}_{i}^{j}$ has length $N_{f}=126$ and describes the frequency content in each unit. Here we use the subscript $i$ to reference specific sound units in the $j$’th waveform. This is itself a linear operation which does not aid in making the data linearly separable.\par

For each sound unit, $\vec{k}_{i}^{j}$ is multiplied by a ‘masking’ matrix $\bf{M}$, which is a $N_{\theta}\times N_{f}$ matrix with binary entries randomly chosen from [-1,1]. The purpose of this ‘masking’ process is to mix all the frequency components together and create the RC input vector  $\vec{x}_{i}^{j}=\bf{M}$$\vec{k}_{i}^{j}$. For this task we choose to emulate $N_{\theta}=100$ ‘virtual’ neurons. The resulting input vector is normalized to a maximum absolute value of 1 which is later mapped to the input voltage range of the RC. \par

The input vectors for each sound unit are concatenated and injected into the RC as a sequence of control voltages applied to the microwave switch. Since the vector $\vec{x}_{i}^{j}$ has been normalized, the values fall in the range [-1,1] which is mapped to the input control voltage range $V_{in}$=[-25 mV,125 mV].  As with the STM and PC tasks, each neuron value is injected to the RC for a fixed input time interval $\theta^{int}$. For this task, linear separability for different inputs is more important than a ‘fading’ memory of past inputs, since each sound unit is treated individually. Thus the nonlinearity of the RC is the more important characteristic for higher accuracy in this task. We choose an input time interval of $\theta^{int}=1.95$ $\mu$s and a feedback gain of $G=0.49$ dB, which results in an adequate STM capacity of 2.14 and maximizes the PC capacity to 1.47. Using this input time interval, each sound unit is processed by the active ring RC in 195 $\mu$s and so entire digits are processed on the order of milliseconds.\par

\begin{figure*}[!t]
	\centering
	\includegraphics[width=1\textwidth]{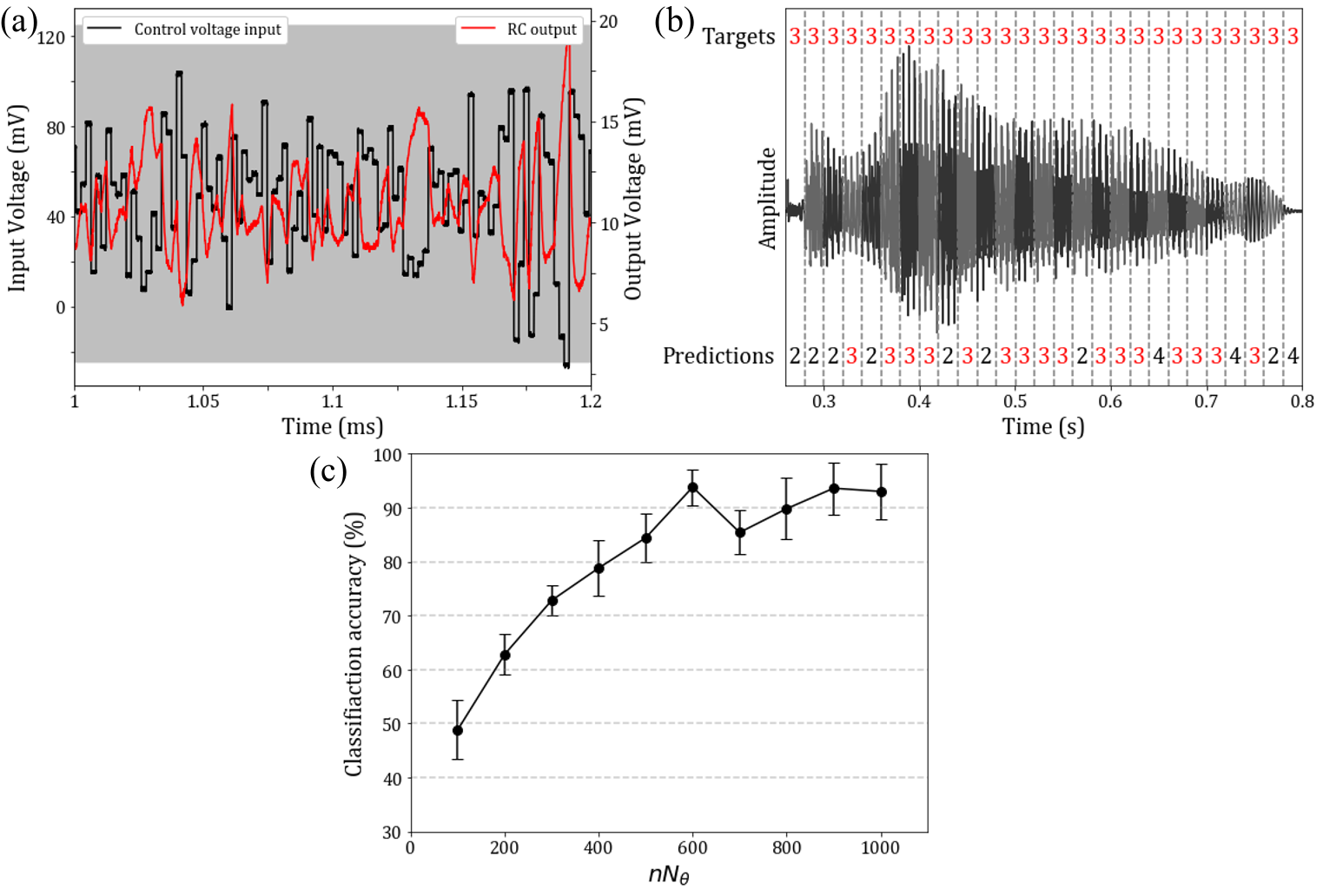}
	\hfil
	\caption{(a) Example of the RC input voltage (black) and the active ring diode output voltage (red). The grey shaded region indicates the computing range of input control voltage. (b) An example of a raw digit waveform sample in the case of a ‘3’. The vertical dashed lines indicate the sound units which the waveform is segmented into. The numbers along the horizontal axis show the predicted digit for each interval. (c) Classification accuracy averaged across the ten repetitions of the training/testing procedure against the total dimensionality of the output vector, $nN_{\theta}$. Error bars show the standard deviation of the classification accuracies.}
	\label{fig:FIG3}
\end{figure*}

Fig. \ref{fig:FIG3}(a) shows an example of the input signal and the active-ring diode output voltage. The input interval time chosen is less than the relaxation time of the active ring, and so the ring signal does not come to equilibrium before the next neuron value is applied. For each input in $\vec{x}_{i}^{j}$, the active ring diode voltage is read by an oscilloscope to create an output vector $\vec{v}_{i}^{j}$. Like with the STM and PC tasks, there is the opportunity to increase the dimensionality of the output by taking multiple samples within each input time interval, $\theta^{int}$. We label the number of samples per neuron as $n$. The length of the RC output vector $\vec{v}_{i}^{j}$ for each sound unit is then $nN_{\theta}$ and each input in $\vec{x}_{i}^{j}$ corresponds to $n$ output values in $\vec{v}_{i}^{j}$.\par

Once all 500 digit samples have been passed through the RC, the last step in the process is training the output weights to reconstruct the targets. This step is done on a computer using simple linear regression techniques. The predictions are reconstructed from the RC outputs by $\vec{y}^{j}=\bf{W}^{out}$$\vec{v}_{i}^{j}$, where the matrix $\bf{W}^{out}$ has dimensions $10 \times nN_{\theta}$  and converts the $nN_{\theta}$  outputs from each sound unit into a column vector of length 10 where each entry corresponds to a specific digit. The optimum weight matrix is that which maps $\vec{v}_{i}^{j}$ exactly onto the correct target vector $\hat{y}^{j}$, which has a value of 1 where the index corresponds to the target digit and 0 otherwise. All sound units in a specific digit are allocated the same target vector as shown by the target values in Fig. \ref{fig:FIG3}(b).\par

Of the ten utterances of each digit spoken by each speaker, we use 9 utterances for training the output weight matrix and use the remaining utterance for testing. Thus our training set has 450 digits and our testing set has 50 digits. Recall that each digit has a further $N_{int}^{j}$ individual sound units. Classification accuracy is chosen as the metric to evaluate performance of our RC concept on this task, which is simply the percentage of the samples in the testing set correctly classified. Since there are 10 different ways to select the 9 utterances used for training and the one utterance used for testing, we repeat the training and testing procedure for all combinations and average the classification accuracies. In this way every sample in the dataset is used in the testing process exactly once.\par

In order to train the weights matrix for all samples in the training set the RC output column vectors $\vec{v}_{i}^{j}$ are concatenated into one matrix $\bf{V}$ with dimensions $nN_{\theta} \times N_{train}$, where $N_{train}$ is the number of training samples. Similarly the target vectors are concatenated into one matrix $\bf{\hat{Y}}$ with dimensions $10 \times N_{train}$. Then the ideal set of output weights satisfies $\bf{\hat{Y}}=\bf{W^{out}}\bf{V}$ and the optimum output matrix is obtained by computing $\bf{W^{out}}=\bf{\hat{Y}} \bf{V}^{-1}$, where we use the Moore-Penrose pseudo-inverse of the matrix $\bf{V}$. This singular training step is very fast. \par

Once the weight matrix has been obtained, the model is evaluated on the test set. In this case we calculate the reconstructed output matrix as: $\bf{Y'^{j}}=\bf{W^{out}}\bf{V'^{j}}$, for each of the 50 digits in the test set. Here the dashes indicate samples contained in the test set. This calculation results in a predicted output column vector for each sound unit in the $j$’th digit. Ideally the column vector has a value of 1 where the index corresponds to the correct digit and 0 otherwise. In reality, the digit which is selected as the prediction corresponds to the vector entry with value closest to 1. So a prediction is made for each sound unit in the digit sample as shown in Fig. \ref{fig:FIG3}(b) for an example of a waveform corresponding to a ‘3’. For the most part, the models correctly classify the sounds units, with some intervals being confused for ‘2’s. To determine the final classification of the waveform, the matrix $\bf{Y'^{j}}$ is averaged across its rows to obtain the final predicted output column vector. The highest value then is the winner. In the case of Fig. \ref{fig:FIG3}(b), the waveform is correctly classified as a ‘3’.\par

Shown in Fig. \ref{fig:FIG3}(c) are the averaged classification accuracies plotted against the length of the RC output vector $\vec{v}_{i}^{j}$. As the effective dimensionality increases, so too does the classification accuracy until it saturates at approximately 93$\%$. \par

In order to assess the RC performance on this task, we carry out a baseline measurement where the output weights matrix is obtained directly from the input vector $\vec{x}_{i}^{j}$. The training procedure remains the same but the RC diode output training matrix $\bf{V}$ is substituted with a matrix of the combined inputs $\bf{X}$. The classification accuracy obtained on the pre-processed data is 25.2$\%$$\pm$ 3.7$\%$. Considering a model which simply guesses which digit the waveform corresponds to would be correct 10$\%$ of the time, the result of 25$\%$ indicates that the pre-processing method used here does help improve the separability of the original data, however not by any significant amount. Making use of the active ring RC can improve the classification accuracy by up to 70$\%$ depending on the dimensionality of the reservoir. Our results are similar to that found in Ref. \cite{torrejon_neuromorphic_2017} for spin-torque nano-oscillators, where for a single node RC with 400 ‘virtual’ neurons, classification accuracies of 80$\%$ were obtained on the same dataset using the simple spectrographic pre-processing method. From Fig. \ref{fig:FIG3}(c) one sees that the equivalently sized reservoir obtains the same 80$\%$ classification accuracy. This is not surprising, while the two concepts vary significantly in the physical operation and design, they share the same RC structure: a single node whose state varies slowly to a given input. \par

\section{Conclusion}
In this letter we evaluated the performance of the spin-wave delay-line active-ring reservoir computer on three baseline tasks. Performance on STM and PC tasks showed that enhancements to the experimental setup have resulted in quantitative improvements on the fading memory and nonlinear characteristics of the RC. Furthermore, this simple RC architecture performs effectively on the spoken digit recognition task, with classification accuracies reaching $93\%$, showing an improvement of $70\%$ from the pre-processed spectrographic data.

\section{Acknowledgments}
Research Collaboration Award and Vice Chancellor’s Senior Research Award from the University of Western Australia are acknowledged. The work of S. Watt was supported by the Australian Government Research Training Program.

\bibliographystyle{IEEEtran}
\bibliography{Digit_references}

\end{document}